# UWB Propagation Characteristics of Human-to-Robot Communication in Automated Collaborative Warehouse

Branimir Ivsic*
Ericsson Nikola Tesla d.d., Research and Development Centre, Radio Development Unit
Zagreb, Croatia
branimir.ivsic@ericsson.com

Juraj Bartolic, Zvonimir Sipus
University of Zagreb,
Faculty of electrical engineering and computing
Zagreb, Croatia
juraj.bartolic@fer.hr, zvonimir.sipus@fer.hr

Josip Babic
Končar–Electrical Engineering Institute Inc.
Zagreb, Croatia
jbabic@koncar-institut.hr

*Abstract*—Propagation of UWB Gaussian signal in a model of automated collaborative warehouse is analyzed using ray tracing method. The transmitting antenna is placed on the human body while the received power profiles in warehouse containing empty racks and racks filled with different loads are calculated. This gives rise to estimation of safe communication range between humans and robots.

## I. INTRODUCTION

One of the most prospective automation concepts for large warehouses is based on the idea of having a fleet of wheeled mobile robots capable of picking and carrying modular portable racks. Nevertheless, since humans still need to be capable to enter the warehouse (e.g. for service, maintenance or some intervention), safety issues pose a problem due to risk of collisions between humans and robots [1 – 3]. The safety concept, based on continuous communication between human and robots, is proposed in [3]. In this scenario, the human is supposed to wear special clothing equiped with antenna and ultrawideband transceiver, while the robots used for carrying warehouse racks need to stop if they reach the safety distance to the human of 6m (Fig. 1). In this paper we explore this concept and perform the modeling and analysis of wave propagation in collaborative warehouse environment in order to estimate the safe communication distance between humans and robots.

In [4] we have proposed several scenarios of an idealized warehouse environment containing PEC rack clusters and have shown that (depending on antenna polarization and surface roughness) it is generally possible to ensure safe communication at distances of over 20 meters by relying basically on reflection and diffraction from rack surfaces and edges. To extend this model we propose here a more realistic warehouse model with randomly distributed and vertically stratified racks containing various items and analyze how various materials (loads) placed at the racks affect signal coverage across the warehouse.

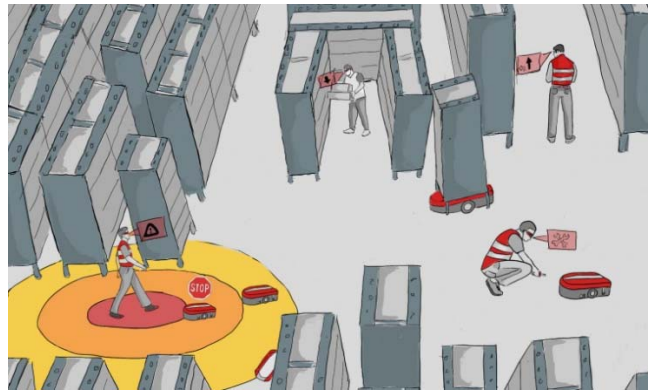

Fig. 1. Human –robot interaction in automated collaborative warehouse [3]

## II. THE WAREHOUSE MODEL

In the proposed model, the communication between the human and robots is realized via an UWB Gaussian pulse centered at 3.994 GHz and 468 MHz wide, which corresponds to UWB Channel 2 [3]. The UWB signal is suitable for multipath environment due to immunity to fading [5]. The antennas to be used are planar circular monopoles with dipole-like behavior (more details are given in [4]). The transmitting antenna is placed at the height of 1.5 m (corresponding to human chest) while the received power profile is calculated at the height of 0.2 m (corresponding to robots height). The input power to the antenna is normalized to 0 dBm and the allowed path loss is 90 dB. For calculations we use ray-tracing software Remcom Wireless Insite [6].

In Fig. 2 we show side view of single vertically stratified rack. It contains four layers of PEC plates (2 cm thick) covered with some item and having a 10 cm air gap above it (thus items

This work has been supported from the European Union's Horizon 2020 research and innovation program under grant agreement No. 688117 (SafeLog), and by Ericsson Nikola Tesla d.d. and University of Zagreb, Faculty of Electrical Engineering and Computing, under the project Emerging Wireless and Information Technologies for 5G Radio Access Networks (EWITA).

do not occupy whole rack floor). The racks are arranged into four clusters (Fig. 3) with 1.5 m wide corridor between each two clusters (each cluster contains seven randomly distributed racks). There is also an additional 5 cm wide vertical air gap between each two neighboring racks. The actual warehouse area is 22 m × 8 m which roughly conforms to the actual testing site [4]. In addition, we assume that the rack surfaces (walls) possess roughness correction (i.e. deviations from flat surface) of $\Delta h$=5cm [7], which is realistic in warehouses due to uneven distribution of items. The transmitting antenna is placed in the middle of the warehouse while all the antennas are vertically polarized (we have shown in [4] that vertical polarization ensures optimum signal coverage in warehouse environment).

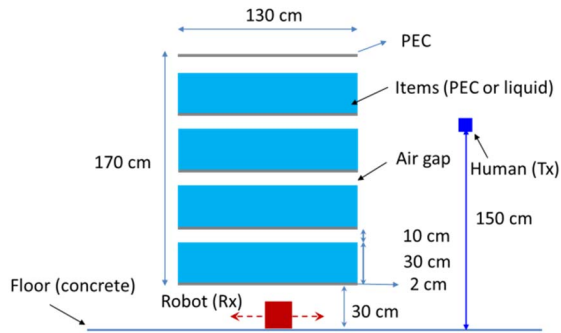

Fig. 2.  Side view of a single stratified rack.

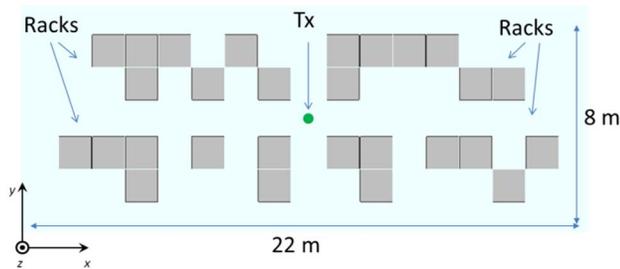

Fig. 3.  Top view of warehouse with sparse stratified racks

### III.  RESULTS AND DISCUSSION

In Figs 4-6 we illustrate the calculated received power profiles for three representative cases. The observed cases are: empty racks, racks filled with olive oil load ($\varepsilon_r$ = 2.87; $\sigma$ = 0.0289), and racks filled with Coca Cola load ($\varepsilon_r$ = 71.25; $\sigma$ = 4.1991) [8]. It can be seen that when the racks are empty (Fig. 4) an excellent signal coverage is obtained within the observation area, while in the cases of olive oil and Coca Cola (Figs 5 and 6) some degradations occur behind the racks (i.e. there are areas with received power less than -90 dBm). Note however that coverage across the corridor where humans walk is still very good. Furthermore, note that the cases of olive oil and Coca Cola in fact possess similar performance despite differences in electrical properties, which means that material put on racks has not major influence onto signal coverage.

This effect is to be theoretically evaluated in the future, together with laboratory measurement setup for evaluation of results.

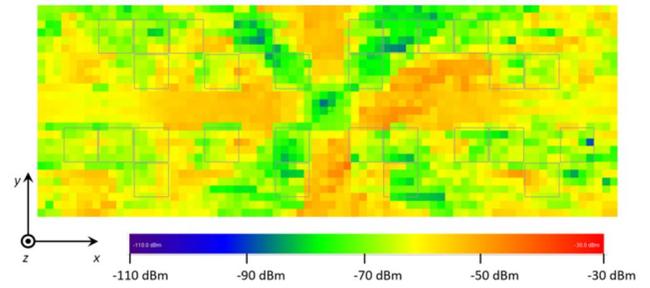

Fig. 4.  Received power profile - empty racks

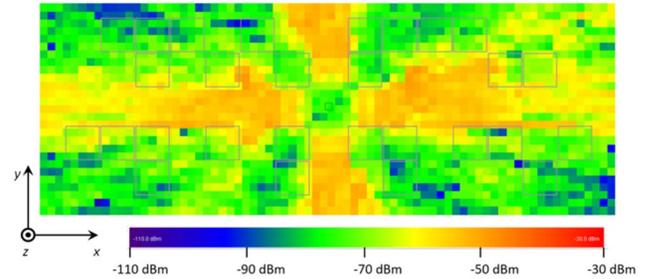

Fig. 5.  Received power profile - racks filled with olive oil

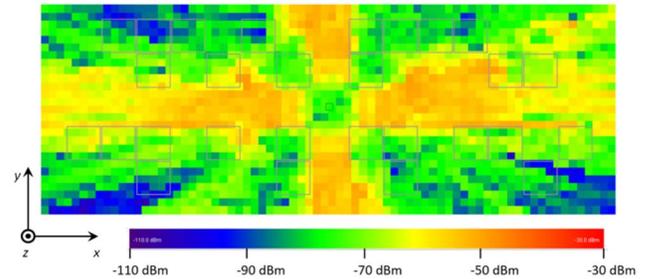

Fig. 6.  Received power profile – racks filled with Coca Cola

REFERENCES

[1] CarryPick: Flexible and modular storage and order picking system. [Online] Available: https://www.swisslog.com/en-us/warehouse-logistics-distribution-center-automation/products-systems-solutions

[2] T. Petković, D. Puljiz, I. Marković, B. Hein, "Human intention estimation based on hidden Markov model motion validation for safe flexible robotized warehouses," Robotics and Computer-Integrated Manufacturing, vol. 57, pp. 182-196, 2019.

[3] Safelog project webpage. (2019) [Online] Available: http://safelog-project.eu/index.php

[4] B. Ivšić, Z. Šipuš, J. Bartolić, J. Babić, "Analysis of Safe Ultrawideband Human-Robot Communication in Automated Collaborative Warehouse", EuCAP 2020, accepted for publication.

[5] S. Sangodoyin et al, "Statistical Modeling of Ultrawideband MIMO Propagation Channel in a Warehouse Environment", IEEE Transactions on Antennas and Propagation, Vol. 64, No. 9, pp. 4049-4062, 2016

[6] Remcom Wireless Insite webpage. (2019) [Online] Available: https://www.remcom.com/wireless-insite-em-propagation-software

[7] L. Boithias, "Radio Wave Propagation", New York: McGraw-Hill, 1987.

[8] A. Dhekne et al., "LiquID: A wireless liquid identifier", in Proceedings of the 16th ACM International Conference on Mobile Systems, Applications, and Services, pp.442-454, Munich (Germany), 2018